\pdfoutput=1

\documentclass[11pt]{article}
\usepackage{nodalida2021}
\usepackage{times}
\usepackage{url}
\usepackage{latexsym}
\usepackage{appendix}
\usepackage{xcolor}
\usepackage{booktabs}
\usepackage{multirow}
\usepackage{hyperref}
\usepackage{adjustbox}
\usepackage{changes}

\aclfinalcopy 
\def\aclpaperid{13} 

\newcommand{\dataset}{\textsc{JobStack}} 
\newcommand{\bertb}{BERT\textsubscript{base}}
\newcommand{\berto}{BERT\textsubscript{Overflow}}
\newcommand{\bertl}{BERT\textsubscript{large} }

\title{De-identification of Privacy-related Entities in Job Postings}

 \author{Kristian Nørgaard Jensen,\textsuperscript{$\diamondsuit$} Mike Zhang\textsuperscript{$\diamondsuit$} and Barbara Plank  \\
  Department of Computer Science\\
  ITU Copenhagen, Denmark \\
  {\tt krnj@itu.dk, mikz@itu.dk, bplank@itu.dk} \\}
\date{}

\begin{document}
\maketitle 
\begingroup\renewcommand\thefootnote{$\diamondsuit$}
\footnotetext{The authors contributed equally to this work.}
\endgroup
\begin{abstract}
De-identification is the task of detecting privacy-related entities in text, such as person names, emails and contact data. It has been well-studied within the medical domain. The need for de-identification technology is increasing, as privacy-preserving data handling is in high demand in many domains. In this paper, we focus on job postings.  We present \dataset, a new corpus for de-identification of personal data in job vacancies on Stackoverflow. We introduce baselines, comparing Long-Short Term Memory (LSTM) and Transformer models. To improve upon these baselines, we experiment with contextualized embeddings and distantly related auxiliary data via multi-task learning. Our results show that auxiliary data improves de-identification performance. 
Surprisingly, vanilla BERT turned out to be more effective than a BERT model trained on other portions of Stackoverflow.
\end{abstract}

\section{Introduction}
It is becoming increasingly important to anonymize privacy-related information in text, such as person names and contact details. The task of de-identification is concerned with detecting and anononymizing such information. Traditionally, this problem has been studied in the medical domain by e.g.,~\citet{szarvas2007state,friedrich-etal-2019-adversarial, trienes2020comparing} to anonymize (or pseudo-anonymize) person-identifiable information in electronic health records (EHR).  With new privacy-regulations (\hyperref[sec:relwork]{Section 2}) de-identification is becoming more important for broader types of text. For example, a company or public institution might seek to de-identify documents before sharing them. On another line, de-identification can benefit society and technology at scale.  Particularly auto-regressive models trained on massive text collections pose a potential risk for exposing private or sensitive information~\cite{carlini2019secret, carlini2020extracting}, and de-identification can be one way to address this.

In this paper, we analyze how effectively sequence labeling models are in identifying privacy-related entities in job posts. To the best of our knowledge, we are the first study that investigates de-identification methods applied to job vacancies. In particular, we examine: How do Transformer-based models compare to LSTM-based models on this task (\textbf{\textsc{RQ1}})?  How does BERT compare to \berto{}~\cite{tabassum2020code} (\textbf{\textsc{RQ2}})? To what extent can we use existing medical de-identification data and Named Entity Recognition (NER) data to improve de-identification performance (\textbf{\textsc{RQ3}})? To answer these questions, we put forth a new corpus, \dataset,  annotated with around 22,000 sentences in English job postings from Stackoverflow for person names, contact details, locations, and information about the profession of the job post itself.




\paragraph{Contributions} We present \dataset, the first job postings dataset with professional and personal entity annotations from  Stackoverflow. Our experiments on entity de-identification with neural methods show that Transformers outperform bi-LSTMs, but surprisingly a BERT variant trained on another portion of Stackoverflow is  less effective. We find auxiliary tasks from both news and the medical domain to help boost performance.

\section{Related Work}\label{sec:relwork}

\subsection{De-identification in the Medical Domain}
De-identification has mostly been investigated in the medical domain (e.g.,~\citet{szarvas2007state, meystre2010automatic, liu2015automatic, jiang2017identification, friedrich-etal-2019-adversarial, trienes2020comparing}) to ensure the privacy of a patient in the analysis of their medical health records. 
Apart from an ethical standpoint, it is also a legal requirement imposed by multiple legislations such as the US Health Insurance Portability and Accountability Act (HIPAA)~\cite{act1996health} and the European General Data Protection Regulation (GDPR)~\cite{regulation2016regulation}. 

Many prior works in the medical domain used the I2B2/UTHealth dataset~\cite{stubbs_annotating_2015} to evaluate de-identification. The dataset consists of clinical narratives, which are free-form medical texts written as a first person account by a clinician. Each of the documents describes a certain event, consultation or hospitalization. All of the texts have been annotated with a set of Protected Health Information (PHI) tags (e.g.\ name, profession, location, age, date, contact, IDs) and subsequently replaced by realistic surrogates.  The dataset was originally developed for use in a shared task for automated de-identification systems. Systems tend to perform very well on this set, in the shared task three out of ten systems achieved F1 scores above 90 \cite{stubbs2015automated}. More recently, systems reach over 98 F1 with neural models~\cite{dernoncourt2017identification, liu2017identification, khin2018deep, trienes2020comparing, johnson2020deidentification}. We took I2B2 as inspiration for annotation of \dataset.

Past methods for de-identification in the medical domain can be categorised in three categories. (1) Rule-based approaches, (2) traditional machine learning (ML)-based systems (e.g., feature-based Conditional Random Fields (CRFs)~\cite{lafferty2001conditional}, ensemble combining CRF and rules, data augmentation, clustering), and (3) neural-based approaches.

\paragraph{Rule-based} First, \citet{gupta2004evaluation} made use of a set of rules, dictionaries, and fuzzy string matching to identify protected health information (PHI). In a similar fashion,~\citet{neamatullah2008automated} used lexical look-up tables, regular expressions, and heuristics to find instances of PHI.

\paragraph{Traditional ML} Second, classical ML approaches employ feature-based 
CRFs~\cite{aberdeen2010mitre, he2015crfs}. Moreover, earlier work showed the use of CRFs in an ensemble with rules~\cite{stubbs2015automated}. Other ML approaches include data augmentation by~\citet{mcmurry2013improved}, where they added public medical texts to properly distinguish common medical words and phrases from PHI and trained decision trees on the augmented data.

\paragraph{Neural methods} Third, regarding neural methods,~\citet{dernoncourt2017identification} were the first to use Bi-LSTMs, which they used in combination with character-level embeddings. Similarly,~\citet{khin2018deep} performed 
de-identification by using a Bi-LSTM-CRF architecture with ELMo embeddings~\cite{peters2018deep}. 
~\citet{liu2017identification} used four individual methods (CRF-based, Bi-LSTM, Bi-LSTM with features, and rule-based methods) for de-identification, and used an ensemble learning method to combine all PHI instances predicted by the three methods.
~\citet{trienes2020comparing} opted for a Bi-LSTM-CRF as well, but applied it with contextual string embeddings~\cite{akbik2018contextual}. 
Most recently,~\citet{johnson2020deidentification} fine-tuned \bertb{} and \bertl{}~\cite{devlin2019bert} for de-identification. Next to ``vanilla'' BERT, they experiment with fine-tuning different domain specific pre-trained language models, such as SciBERT~\cite{beltagy2019scibert} and BioBERT~\cite{lee2020biobert}. They achieve state-of-the art performance in de-identification on the I2B2 dataset with the fine-tuned \bertl{} model. 
From a different perspective, the approach of ~\citet{friedrich-etal-2019-adversarial} is based on adversarial learning, which automatically pseudo-anonymizes EHRs.

\begin{table*}[ht!]
\centering
\resizebox{0.65\linewidth}{!}{
\begin{tabular}{@{}lrrrr@{}}
\toprule
                                                & Train             & Dev               & Test             & Total\\\midrule
\multicolumn{1}{l|}{Time}                       & June -- August 2020  & \multicolumn{2}{c}{September 2020}  & \multicolumn{1}{|r}{-} \\\midrule
\multicolumn{1}{l|}{\# Documents}               & 313                  & 41                & 41               & \multicolumn{1}{|r}{395} \\
\multicolumn{1}{l|}{\# Sentences}               & 18,055             & 2082              & 2092             & \multicolumn{1}{|r}{22,219} \\
\multicolumn{1}{l|}{\# Tokens}                  & 195,425            & 22,049             & 21,579            & \multicolumn{1}{|r}{239,053}  \\
\multicolumn{1}{l|}{\# Entities}                & 4,057              & 462               & 426              & \multicolumn{1}{|r}{5,154} \\\midrule

\multicolumn{1}{l|}{avg. \# sentences}          & 57.68             & 50.78             & 51.02            & \multicolumn{1}{|r}{53.16} \\ 
\multicolumn{1}{l|}{avg. tokens / sent.}        & 10.82             & 10.59             & 10.32            & \multicolumn{1}{|r}{10.78}\\ 
\multicolumn{1}{l|}{avg. entities / sent.}      & 0.22              & 0.22              & 0.20             & \multicolumn{1}{|r}{0.21} \\ 
\multicolumn{1}{l|}{density}                    & 14.73             & 14.31             & 14.58            & \multicolumn{1}{|r}{14.54} \\\midrule

\multicolumn{1}{l|}{\texttt{Organization}}      & 1803              & 215               & 208               & \multicolumn{1}{|r}{2226}\\
\multicolumn{1}{l|}{\texttt{Location}}          & 1511              & 157               & 142               & \multicolumn{1}{|r}{1810}\\
\multicolumn{1}{l|}{\texttt{Profession}}        & 558               & 63                & 64                & \multicolumn{1}{|r}{685}\\
\multicolumn{1}{l|}{\texttt{Contact}}           & 99               & 10                & 7                & \multicolumn{1}{|r}{116}\\
\multicolumn{1}{l|}{\texttt{Name}}              & 86                & 17                & 5                 & \multicolumn{1}{|r}{108}\\
\bottomrule
\end{tabular}
}
\caption{Statistics of our \dataset{}  dataset.}
\label{tab:datastats}
\end{table*}

\subsection{De-identification in other Domains}
Data protection in general however is not only limited to the medical domain. Even though work outside the clinical domain is rare, personal and sensitive data is in abundance in all kinds of data. For example, \citet{eder2019identification} pseudonymised German emails.~\citet{bevendorff2020crawling} published a large preprocessed email corpus, where only the email addresses themselves where anonymized.
Apart from emails, several works went into de-identification of SMS messages~\cite{treurniet2012collection, patel2013approaches, chen2013creating} in Dutch, French, English and Mandarin respectively. Both ~\citet{treurniet2012collection, chen2013creating} conducted the same strategy and automatically anonymized all occurrences of dates, times, decimal amounts, and numbers with more than one digit (telephone numbers, bank accounts, et cetera), email addresses, URLs, and IP addresses. All sensitive information was replaced with a placeholder.~\citet{patel2013approaches} introduced a system to anonymize SMS messages by using dictionaries. It uses a dictionary of first names and  anti-dictionaries (of ordinary language and of some forms of SMS writing) to identify the words that require anonymization.

In our work, we study de-identification for names, contact information, addresses, and professions, as further described in \hyperref[sec:dataset]{Section 3}.

\section{\dataset{} Dataset}\label{sec:dataset}
\begin{figure*}
    \centering
\noindent\fbox{%
    \parbox{\textwidth}{%
    \small
...\\
13. Job description:\\ 
14. \textbf{\lbrack XXX\textsubscript{Organization}\rbrack} is a modern multi tenant, microservices based solution and Floor Planning is one major functional solution vertical of the \textbf{\lbrack XXX\textsubscript{Organization}\rbrack} platform. \\

15. What you’ll be doing:\\
16. As a \textbf{\lbrack XXX\textsubscript{Profession}\rbrack} for \textbf{\lbrack XXX\textsubscript{Organization}\rbrack}, you will be one of the founding members of our \textbf{\lbrack XXX\textsubscript{Location}\rbrack} based floor planning development team.\\
17. You will be in charge for development of future floor planning capabilities on the \textbf{\lbrack XXX\textsubscript{Organization}\rbrack} platform and be the software architect for the capability.\\
18. You will drive the team to improve the coding practices and boost performance.\\ 
19. You will also be a member of our \textbf{\lbrack XXX\textsubscript{Organization}\rbrack} and have a major influence on feature roadmap and technologies we use.\\
...
        }%
}
    \caption{Snippet of a job posting, full job posting can be found in \hyperref[sec:jobpost]{Appendix A}.}
    \label{fig:snippet}
\end{figure*}

In this section, we describe the \dataset{} dataset.
There are two basic approaches to remove privacy-bearing data from the job postings.
First, anonymization identifies instances of personal data (e.g. names, email addresses, phone numbers) and replaces these strings by some placeholder (e.g.\ \texttt{\{name\}}, \texttt{\{email\}}, \texttt{\{phone\}}). 
The second approach, pseudonymisation preserves the information of personal data by replacing these privacy-bearing strings with randomly chosen alternative strings from the same privacy type (e.g. replacing a name with ``John Doe''). 
The term de-identification subsumes both anonymization and pseudonymisation. In this work, we focus on anonymization.\footnote{\citet{meystre2015identification} notes that de-identification means removing or replacing personal identifiers to make it difficult to reestablish a link between the individual and his or her data, but it does not make this link impossible. 
}

\citet{eder2019identification} argues that the anonymization approach might be appropriate to eliminate privacy-bearing data in the medical domain, but would be inappropriate for most Natural Language Processing (NLP) applications since crucial discriminative information and contextual clues will be erased by anonymization.

If we shift towards pseudonymisation, we argue that there is still the possibility to resurface the original personal data. Henceforth, our goal is to anonimize job postings to the extent that one would not be able to easily identify a company from the job posting.
However, as job postings are public, we are aware that it would be simple to find the original company that posted it with a search engine.
Nevertheless, we abide to the GDPR compliance which requires us to protect the personal data and privacy of EU citizens for transactions that occur within EU member states \cite{regulation2016regulation}. In job postings this would be the names of employees, and their corresponding contact information.\footnote{\url{https://ec.europa.eu/info/law/law-topic/data-protection/reform/rules-business-and-organisations/application-regulation/do-data-protection-rules-apply-data-about-company_en}}

Over a period of time, we scraped 2,755 job postings from Stackoverflow and selected 395 documents to annotate, the subset ranges from June 2020 to September 2020.
We manually annotated the job postings with the following five entities: \texttt{Organization}, \texttt{Location}, \texttt{Contact}, \texttt{Name}, and \texttt{Profession}. 

To make the task as realistic as possible, we kept all sentences in the documents. The statistics provided in the following therefore reflect the natural distribution of entities in the data.
A snippet of an example job post can be seen in \hyperref[fig:snippet]{Figure 1}, the full job posting can be found in \hyperref[sec:jobpost]{Appendix A}.

\subsection{Statistics}

\hyperref[tab:datastats]{Table 1} shows the statistics of our dataset.
We split our data in 80\% train, 10\% development, and 10\% test. 
Besides of a regular document-level random split, ours is further motivated based on time.
The training set covers the job posts posted between June to August 2020 and the development- and test set are posted in September 2020.
To split the text into sentences, we use the \texttt{sentence-splitter} library used for processing the Europarl corpus \cite{koehn2005europarl}.
In the training set, we see that the average number of sentences is higher than in the development- and test set (6-7 more). We therefore also calculate the density of the entities, meaning the percentage of sentences with at least one entity.
The table shows that 14.5\% of the sentences in \dataset{} contain at least one entity. Note that albeit having 
document boundaries, we treat the task of de-identification as a standard word-level sequence labeling task.

\subsection{Annotation Schema}
\label{sec:schema}
The aforementioned entity tags are based on the English I2B2/UTHealth corpus \cite{stubbs_annotating_2015}. The tags are more coarse-grained than the I2B2 tags. For example, we do not distinguish between zip code and city, but tag them with \texttt{Location}. We give a brief explanation of the tags.\\
\texttt{Organization}: This includes all companies and their legal entity mentioned in the job postings. The tag is not limited to the company that authored the job posting, but does also include mentions of stakeholders or any other company.\\
\texttt{Location}: This is the address of the company in the job posting. The location also refers to all other addresses, zip codes, cities, regions, and countries mentioned throughout the text. This is not limited to the company  address, but should be used for all location names in the job posting, including abbreviations.\\
\texttt{Contact}: The label includes, URLs, email addresses and phone numbers. This could be, but is not limited to, contact info of an employee from the authoring company.\\
\texttt{Name}: This label covers names of people. This could be, but is not limited  to, a person from the company, such as the contact person, CEO, or the manager. All names appearing in the job posting should be annotated no matter the relation to the job posting itself. Titles such as Dr.\ are not part of the annotation. Apart from people names in our domain, difficulties could arise with other type of names. An example would be project names, with which one could identify a company. In this work, we did not annotate such names.\\
\texttt{Profession}: This label covers the profession that is being searched for in the job posting or desired prior relevant jobs for the current profession. We do not annotate additional meta information such as gender (e.g. Software Engineer (f/m)). We also do not annotate mentions of colleague positions in neither singular or plural form. For example: ``\textit{As a Software Engineer, you are going to work with Security Engineers}''. Here we annotate Software Engineer as profession, but we do not annotate Security Engineers. While this may sound straightforward, however, there are difficulties in regards to annotating professions. A job posting is free text, meaning that one can write anything they prefer to make the job posting as clear as possible (e.g., \textit{Software Engineer (at a unicorn start-up based in [..]}). The opposite is also possible, when they are looking for one applicant to fill in one of multiple positions. For example, ``\textit{We are looking for an applicant to fill in the position of DevOps/Software Engineer}''. From our interpretation, they either want a ``DevOps Engineer'' or a ``Software Engineer''. We decided to annotate the full string of characters ``DevOps/Software Engineer'' as a profession.

\subsection{Annotation Quality}




\begin{table}[ht]
\centering
\begin{tabular}{@{}llll@{}}
\toprule
\multicolumn{1}{l}{}                                         & Token    & Entity    & Unlabeled          \\\midrule
\multicolumn{1}{l|}{A1 -- A2}                                & 0.889    & 0.767     & 0.892             \\
\multicolumn{1}{l|}{A1 -- A3}                                & 0.898    & 0.782     & 0.904             \\
\multicolumn{1}{l|}{A2 -- A3}                                & 0.917    & 0.823     & 0.920             \\\midrule
\multicolumn{1}{l|}{Fleiss' $\kappa$}                        & 0.902    & 0.800     & 0.906             \\\bottomrule
\end{tabular}
\caption{Inter-annotator agreement of the annotators. We show agreement over pairs with Cohen's $\kappa$ and all annotators with Fleiss' $\kappa$.}
\label{tab:iaahuman}
\end{table}

\begin{table*}[ht]
\centering
\begin{tabular}{@{}llll@{}}
\toprule
\multicolumn{1}{l}{Model}                     & F1 Score          & Precision         & Recall \\ \midrule
\multicolumn{1}{l|}{Bilty}                    & $71.76 \pm 2.57$  & $79.00 \pm 1.10$  & $65.80 \pm 3.72$  \\
\multicolumn{1}{l|}{Bilty + CRF}              & $75.15 \pm 0.66$  & $84.09 \pm 1.90$  & $67.96 \pm 0.81$ \\
\multicolumn{1}{l|}{Bilty + Glove 50d}        & $72.53 \pm 0.83$  & $79.21 \pm 2.19$  & $67.03 \pm 2.76$ \\
\multicolumn{1}{l|}{Bilty + Glove 50d + CRF}  & $72.74 \pm 2.23$  & $82.93 \pm 0.87$  & $64.93 \pm 3.93$ \\
\multicolumn{1}{l|}{Bilty + \bertb}           & $77.99 \pm 0.91$  & $83.70 \pm 0.58$  & $73.01 \pm 1.34$  \\
\multicolumn{1}{l|}{Bilty + \bertb{} + CRF}   & $80.09 \pm 0.60$  & $\mathbf{88.23 \pm 0.87}$  & $73.30 \pm 1.47$  \\
\multicolumn{1}{l|}{Bilty + \berto}           & $52.01 \pm 3.15$  & $70.86 \pm 0.68$  & $41.27 \pm 4.19$ \\
\multicolumn{1}{l|}{Bilty + \berto{} + CRF}   & $53.08 \pm 2.88$  & $77.79 \pm 1.20$  & $40.33 \pm 2.98$ \\ \midrule
\multicolumn{1}{l|}{MaChAmp + \bertb }        & $85.70 \pm 0.13$  & $86.66 \pm 0.73$  & $84.78 \pm 0.44$ \\
\multicolumn{1}{l|}{MaChAmp + \bertb{} + CRF} & $\mathbf{86.27 \pm 0.31}$  & $86.40 \pm 0.62$   & $\mathbf{86.15 \pm 0.00}$\\
\multicolumn{1}{l|}{MaChAmp + \berto}         & $65.84 \pm 0.48$  & $70.88 \pm 0.17$  & $61.47 \pm 0.81$  \\
\multicolumn{1}{l|}{MaChAmp + \berto{} + CRF} & $69.35 \pm 0.96$  & $77.27 \pm 3.68$  & $63.06 \pm 2.11$  \\ 
\bottomrule
\end{tabular}
\label{tab:ourdata}
\caption{Results on the development set across three runs using our \dataset{} dataset.}
\end{table*}

To evaluate our annotation guidelines, a sample of the data was annotated by three annotators, one with a background in Linguistics (A1) and two with a background in Computer Science (A2, A3). 
We used an open source text annotation tool named \texttt{Doccano} \cite{doccano}. 
There are around 1,500 overlapping sentences that we calculated agreement on. 
The annotations were compared using Cohen's $\kappa$~\cite{fleiss1973equivalence} between pairs of annotators, and Fleiss' $\kappa$~\cite{fleiss1971measuring}, which generalises Cohen's $\kappa$ to more than two concurrent annotations. 
\hyperref[tab:iaahuman]{Table 2} shows three levels of $\kappa$ calculations, we follow~\citet{balasuriya2009named}'s approach of calculating agreement in NER. (1) \texttt{Token} is calculated on the token level, comparing the agreement of annotators on each token (including non-entities) in the annotated dataset. 
(2) \texttt{Entity} is calculated on the agreement between named entities alone, excluding agreement in cases where all annotators agreed that a token was not a named-entity. 
(3) \texttt{Unlabeled} refers to the agreement between annotators on the exact span match over the surface string, regardless of the type of named entity (i.e., we only check the position of tag without regarding  the type of the named entity).
\citet{landis1977measurement} state that a $\kappa$ value greater than 0.81 indicates almost perfect agreement. 
Given this, all annotators are in strong agreement. 

After this annotation quality estimation, we finalized the guidelines. They formed the basis for the professional linguist annotator, who annotated and finalized the entire final \dataset{} dataset.




\begin{table*}[t]
\centering
\resizebox{\linewidth}{!}{
\begin{tabular}{@{}lllll@{}}
\toprule
Model                               & Auxiliary tasks             & F1 Score          & Precision         & Recall            \\ \midrule
\multicolumn{1}{l|}{\multirow{3}{*}{Bilty + \bertb{} + CRF}}      & \dataset{} + CoNLL          & $81.90 \pm 0.32$  & $86.91 \pm 1.94$  & $77.49 \pm 1.87$  \\
\multicolumn{1}{l|}{}                                             & \dataset{} + I2B2           & $79.15 \pm 2.19$  & $83.61 \pm 2.61$  & $75.18 \pm 2.59$  \\
\multicolumn{1}{l|}{}                                             & \dataset{} + CoNLL + I2B2   & $81.37 \pm 2.01$  & $84.92 \pm 1.67$  & $78.28 \pm 4.34$  \\ \midrule
\multicolumn{1}{l|}{\multirow{3}{*}{Bilty + \berto{} + CRF}}      & \dataset{} + CoNLL          & $58.62 \pm 1.46$  & $79.34 \pm 2.34$  & $46.54 \pm 1.99$  \\
\multicolumn{1}{l|}{}                                             & \dataset{} + I2B2           & $55.99 \pm 1.93$  & $72.03 \pm 6.48$  & $46.10 \pm 2.55$  \\
\multicolumn{1}{l|}{}                                             & \dataset{} + CoNLL + I2B2   & $59.15 \pm 2.15$  & $71.20 \pm 4.80$  & $50.86 \pm 3.31$  \\ \midrule
\multicolumn{1}{l|}{\multirow{3}{*}{MaChAmp + \bertb{} + CRF}}    & \dataset{} + CoNLL          & $\mathbf{87.20 \pm 0.34}$  & $87.24 \pm 1.94$  & $\mathbf{87.23 \pm 1.24}$  \\
\multicolumn{1}{l|}{}                                             & \dataset{} + I2B2           & $86.64 \pm 0.53$  & $\mathbf{88.44 \pm 0.84}$  & $84.92 \pm 0.44$  \\
\multicolumn{1}{l|}{}                                             & \dataset{} + CoNLL + I2B2   & $86.06 \pm 0.66$  & $86.13 \pm 0.50$  & $86.00 \pm 0.87$  \\ \midrule
\multicolumn{1}{l|}{\multirow{3}{*}{MaChAmp + \berto{} + CRF}}    & \dataset{} + CoNLL     & $70.62 \pm 0.64$  & $75.65 \pm 1.41$  & $66.24 \pm 0.98$  \\
\multicolumn{1}{l|}{}                                             & \dataset{} + I2B2           & $73.88 \pm 0.16$  & $80.26 \pm 1.32$  & $68.47 \pm 1.03$  \\
\multicolumn{1}{l|}{}                                             & \dataset{} + CoNLL + I2B2   & $73.29 \pm 0.22$  & $77.66 \pm 0.82$  & $69.41 \pm 0.89$  \\
\bottomrule
\end{tabular}
}
\label{tab:mtlresults}
\caption{Performance of multi-task learning on the development set across three runs.}
\end{table*}

\section{Methods}
\label{sec:methods}
For entity de-identification we use a classic Named Entity Recognition (NER) approach using a Bi-LSTM with a CRF layer. On top of this we evaluate the performance of Transformer-based models with two different pre-trained BERT variants. Furthermore, we evaluate the helpfulness of auxiliary tasks, both using data close to our domain, such as de-identification of medical notes, and more general NER, which covers only a subset of the entities. Further details on the data are given in~\hyperref[sec:auxtasks]{Section 4.3}. 

\subsection{Models}

Firstly, we test a Bi-LSTM sequence tagger (Bilty) \cite{plank-etal-2016}, both with and without a CRF layer. The architecture is similar to the widely used models in previous works. For example, preliminary results of Bilty versus~\citet{trienes2020comparing} show accuracy almost identical to each other: 99.62\% versus 99.76\%. Next we test a Transformer based model, namely the MaChAmp \cite{vandergoot-etal-2020-machamp} toolkit. Current research shows good results for NER using a Transformer model without a CRF layer~\cite{martin-etal-2020-camembert}, hence we tested MaChAmp both with and without a CRF layer for predictions. For both models, we use their default parameters.

\subsection{Embeddings}

For embeddings, we tested with no pre-trained embeddings, pre-trained Glove \cite{pennington2014glove} embeddings, and Transformer-based pre-trained embeddings. For Transformer-based embeddings we focused our attention on two BERT models, \bertb{}~\cite{devlin2019bert} and \berto{}~\cite{tabassum2020code}. When using the Transformer-based embeddings with the Bi-LSTM, the embeddings were fixed and did not get updated during training. 

Using the MaChAmp \cite{vandergoot-etal-2020-machamp} toolkit, we fine-tune the BERT variant with a Transformer encoder. For the Bi-LSTM sequence tagger, we first derive BERT representations as input to the tagger. The tagger further uses word and character embeddings which are updated during model training.

The \berto{} model is a transformer  with the same architecture as \bertb{}. It has been trained from scratch on a large corpus of text from the Q\&A section of Stackoverflow, making it closer to our text domain than the ``vanilla'' BERT model. However, \berto{} is not trained on the job postings portion of Stackoverflow.

\subsection{Auxiliary tasks}
\label{sec:auxtasks}
Both the Bi-LSTM \cite{plank-etal-2016} and the MaChAmp \cite{vandergoot-etal-2020-machamp} toolkit are capable of Multi Task Learning (MTL)~\cite{caruana1997multitask}. We therefore, set up a number of experiments testing the impact of three different auxiliary tasks. The auxiliary tasks and their datasets are as follows:

\begin{itemize}
    \item I2B2/UTHealth~\cite{stubbs_annotating_2015} - Medical de-identification;
    \item CoNLL 2003~\cite{sang2003introduction} - News Named Entity Recognition;
    \item The combination of the above.
\end{itemize}

The data of the two tasks are similar to our dataset in two different ways. The I2B2 lies in a different text domain, namely medical notes, however, the label set of the task is close to our label set, as mentioned in \hyperref[sec:schema]{Section 3.2}.
For CoNLL, we have a general corpus of named entities but fewer types (location, organization, person, and miscellaneous), but the text domain is presumably closer to our data. 
We test the impact of using both auxiliary tasks along with our own dataset. 

\begin{table*}[ht]
\centering
\resizebox{\linewidth}{!}{
\begin{tabular}{@{}lllll@{}}
\toprule
Model                                                        & Auxiliary tasks       & F1 Score & Precision & Recall \\ \midrule
\multicolumn{1}{l|}{Bilty + \bertb{} + CRF}                    & \dataset{}              & $78.99 \pm 0.32$ & $\mathbf{82.44 \pm 0.95}$ & $75.90 \pm 1.39$ \\ \midrule
\multicolumn{1}{l|}{\multirow{4}{*}{MaChAmp + \bertb{} + CRF}} & \dataset{}              & $79.91 \pm 0.38$ & $75.92 \pm 0.39$ & $84.35 \pm 0.49$ \\
\multicolumn{1}{l|}{}                                        & \dataset{} + CoNLL & $81.27 \pm 0.28$ & $77.84 \pm 1.19$ & $85.06 \pm 0.91$ \\
\multicolumn{1}{l|}{}                                        & \dataset{} + I2B2       & $\mathbf{82.05 \pm 0.80}$ & $80.30 \pm 0.99$ & $83.88 \pm 0.67$ \\
\multicolumn{1}{l|}{}                                        & \dataset{} + CoNLL + I2B2       & $81.47 \pm 0.43$ & $77.66 \pm 0.58$ & $\mathbf{85.68 \pm 0.57}$ \\ \bottomrule
\end{tabular}
}
\caption{Evaluation of the best performing models on the test set across three runs.}
\label{tab:testscores}
\end{table*}

\section{Evaluation}

Here we will outline the results of the experiments described in \hyperref[sec:methods]{Section 4}. 
All results are mean scores across three different runs.\footnote{We sampled three random seeds: $3477689$, $4213916$, $8749520$ which are used for all experiments.} The metrics are all calculated using the \texttt{conlleval} script\footnote{\url{https://www.clips.uantwerpen.be/conll2000/chunking/output.html}} from the original CoNLL-2000 shared task. \hyperref[tab:ourdata]{Table 3} shows the results from training on \dataset{} only, \hyperref[tab:mtlresults]{Table 4} shows the results of the MTL experiments described in \hyperref[sec:auxtasks]{Section 4.3}. Both report results on the development set. Lastly, \hyperref[tab:testscores]{Table 5} shows the scores from evaluating selected best models as found on the development set, when tested on the final held-out test set. 

\paragraph{Is a CRF layer necessary?}

In \hyperref[tab:ourdata]{Table 3}, as expected, adding the CRF for the Bi-LSTM clearly helps, and consistently improves precision and thereby F1 score. For the stronger BERT model the overall improvement is smaller and does not necessarily stem from higher precision. We note that on average across the three seed runs, MaChAmp with \bertb{} and no CRF mistakenly adds an I-tag following an O-tag $8$ times out of $426$ gold entities. In contrast, the MaChAmp with \bertb{} and CRF, makes no such mistake in any of its three seed runs. 
Earlier research, such as ~\citet{souza2019portuguese} show that BERT models with a CRF layer improve or perform similarly to its simpler variants when comparing the overall F1 scores. Similarly, they note that in most cases it shows higher precision scores but lower recall, as in our results for the development set. However, interestingly, the precision drops during test for the Transformer-based model. As the overall F1 score increases slightly, we use the CRF layer in all subsequent experiments. The main take-away here is that both models benefit from an added CRF layer for the task, but the Transformer model to a smaller degree.

\paragraph{LSTM versus Transformer}
Initially, LSTM networks dominated the de-identification field in the medical domain. Up until recently, large-scale pre-trained language models have been ubiquitous in NLP, although rarely used in this field. On both development and test results (\hyperref[tab:ourdata]{Table 3}, \hyperref[tab:testscores]{Table 5}), we show that a Transformer-based model outperforms the LSTM-based approaches with non-contextualized and contextualized representations.

\paragraph{Poor performance with \berto}
\bertb{} is the best embedding method among all experiments using Bilty, with \berto{} being the worst with a considerable margin. Being able to fine-tune \bertb{} does give a good increase in performance overall. The same trend is apparent with fine-tuning \berto{}, but it is not enough to catch up with \bertb. We see that overall MaChAmp with \bertb{} and CRF is the best model. However, Bilty with \bertb{} and CRF does have the best precision. 

We hypothesized the domain-specific \berto{} representations would be beneficial for this task. 
Intuitively, \berto would help with detecting profession entities.
Profession entities contain specific skills related to the IT domain, such as \textit{Python developer}, \textit{Rust developer}, \textit{Scrum master}.
Although the corpus it is trained on is not one-to-one to our vacancy domain, we expected to see at most a slight performance drop. 
This is not the case, as the drop in performance turned out to be high. It is not fully clear to us why this is the case. It could be the Q\&A data it is trained on consists of more informal dialogue than in job postings. In the future, we would like to compare these results to training a BERT model on job postings data.


\paragraph{Auxiliary data increases performance}
Looking at the results from the auxiliary experiments in~\hyperref[tab:mtlresults]{Table 4} we see that all auxiliary sources are beneficial, for both types of models. 
A closer look reveals that once again MaChAmp with \bertb{} is the best performer across all three auxiliary tasks. Also, we see that Bilty with \bertb{} has good precision, though not the best this time around. For a task like de-identification recall is preferable, thereby showing that fine-tuning BERT is better than the classic Bi-LSTM-CRF. Moreover, we see that \berto{} is under-performing compared to \bertb{}. However, \berto{} is able to get a four point increase in F1 with I2B2 as auxiliary task in MaChAmp. For Bilty with \berto{} we see a slightly greater gain with both CoNLL and I2B2 as auxiliary tasks. When comparing the auxiliary data sources to each other, we note that the closer text domain (CoNLL news) is more beneficial than the closer label set (I2B2) from a more distant medical text source. This is consistent for the strongest models.

In general, it can be challenging to train multi-task networks that outperform or even  match their single-task counterparts~\cite{alonso2017multitask, clark2019bam}.~\citet{ruder2017overview} mentions training on a large number of tasks is known to help regularize multi-task models. A related benefit of MTL is the transfer of learned ``knowledge'' between closely related tasks. In our case, it has been beneficial to add auxiliary tasks to improve our performance on both development and test compared to a single task setting. In particular, it seemed to have helped with pertaining a high recall score.

\paragraph{Performance on the test set}
Finally, we evaluate the best performing models on our held out test set. The best models are selected based on their performance on F1, precision, and recall. The results are seen in \hyperref[tab:testscores]{Table 5}. Comparing the results to those seen in \hyperref[tab:ourdata]{Table 3} and \hyperref[tab:mtlresults]{Table 4} it is clear to see that Bilty with \bertb{} sees a smaller drop in F1 compared to that of MaChAmp with \bertb. We do also see an increase in recall for Bilty compared to its performance on the development set. In general we see that recall for each model is staying quite stable without any significant drops. It is also interesting to see that, the internal ranking between MTL MaChAmp with \bertb{} has changed, with \dataset{} + I2B2 being the best performing model in terms of F1. 

\paragraph{Per-entity Analysis} 

\begin{table}[t]
\centering
\resizebox{\linewidth}{!}{
\begin{tabular}{@{}llrr@{}}
\toprule
            &                                & \multicolumn{2}{c}{MaChAmp + \dataset{}} \\ \cmidrule(l){3-4} 
\multicolumn{1}{l}{\multirow{1}{*}{Entity}}            &                            & \multicolumn{1}{c}{+ CoNLL} & \multicolumn{1}{c}{+ I2B2} \\ \midrule
\multicolumn{1}{l|}{\multirow{3}{*}{\textbf{Organization} (208)}} & \multicolumn{1}{l|}{F1}   & $77.51 \pm 0.81$           & $78.34 \pm 1.32$           \\
\multicolumn{1}{l|}{}                                 & \multicolumn{1}{l|}{P}    & $73.73 \pm 1.66$           & $77.86 \pm 1.60$           \\
\multicolumn{1}{l|}{}                                       & \multicolumn{1}{l|}{R}    & $81.73 \pm 0.96$           & $78.85 \pm 1.74$           \\ \midrule
\multicolumn{1}{l|}{\multirow{3}{*}{\textbf{Location} (142)}}     & \multicolumn{1}{l|}{F1}   & $86.88 \pm 1.51$           & $86.67 \pm 1.80$           \\
\multicolumn{1}{l|}{}                                 & \multicolumn{1}{l|}{P}    & $83.86 \pm 1.82$           & $83.47 \pm 1.19$           \\
\multicolumn{1}{l|}{}                                       & \multicolumn{1}{l|}{R}    & $90.14 \pm 1.41$           & $90.14 \pm 2.54$           \\ \midrule
\multicolumn{1}{l|}{\multirow{3}{*}{\textbf{Profession} (64)}}   & \multicolumn{1}{l|}{F1}   & $80.20 \pm 2.76$           & $83.88 \pm 0.90$           \\
\multicolumn{1}{l|}{}                                  & \multicolumn{1}{l|}{P}    & $77.44 \pm 3.82$           & $82.42 \pm 0.63$           \\
\multicolumn{1}{l|}{}                                       & \multicolumn{1}{l|}{R}    & $83.33 \pm 4.51$           & $85.42 \pm 1.80$           \\ \midrule
\multicolumn{1}{l|}{\multirow{3}{*}{\textbf{Contact} (7)}}      & \multicolumn{1}{l|}{F1}   & $87.91 \pm 3.81$           & $75.48 \pm 4.30$           \\
\multicolumn{1}{l|}{}                                   & \multicolumn{1}{l|}{P}    & $90.47 \pm 8.25$           & $71.03 \pm 4.18$           \\
\multicolumn{1}{l|}{}                                       & \multicolumn{1}{l|}{R}    & $85.71 \pm 0.00$           & $80.95 \pm 8.24$           \\ \midrule
\multicolumn{1}{l|}{\multirow{3}{*}{\textbf{Name} (5)}}         & \multicolumn{1}{l|}{F1}   & $86.25 \pm 8.08$           & $85.86 \pm 4.38$           \\
\multicolumn{1}{l|}{}                                   & \multicolumn{1}{l|}{P}    & $76.39 \pm 12.03$          & $75.40 \pm 6.87$           \\
\multicolumn{1}{l|}{}                                       & \multicolumn{1}{l|}{R}    & $100.00 \pm 0.00$          & $100.00 \pm 0.00$           \\ \bottomrule
\end{tabular}
}
\caption{Performance of the two different auxiliary tasks. Reported is the F1, Precision (P), and Recall (R) per entity. The number behind the entity name is the gold label instances in the test set.}
\label{tab:entityanalysis}
\end{table}

In \hyperref[tab:entityanalysis]{Table 6}, we show a deeper analysis on the test set: the performance of the two different auxiliary tasks in a multi-task learning setting, namely CoNLL and I2B2. We hypothesized different performance gains with each auxiliary task. For I2B2, we expected \texttt{Contact} and \texttt{Profession} to do better than CoNLL, since I2B2 contains contact information entities (e.g., phone numbers, emails, et cetera) and professions of patients. Surprisingly, this is not the case for \texttt{Contact}, as CoNLL outperforms I2B2 on all three metrics. We do note however this result could be due to little instances of \texttt{Contact} and \texttt{Name} being present in the gold test set. Additionally, both named entities are predicted six to nine times by both models on all three runs on the test set. This could indicate the strong difference in performance.
For \texttt{Profession}, it shows that I2B2 is beneficial for this particular named entity as expected. For the other three named entities, the performance is similar. As \texttt{Location}, \texttt{Name}, and \texttt{Organization} are in both datasets, we did not expect any difference in performance. The results confirm this intuition.

\section{Conclusions}
In this work we introduce \dataset{}, a dataset for de-identification of English Stackoverflow job postings. 
Our implementation is publicly available.\footnote{\url{https://github.com/kris927b/JobStack}} The dataset is freely available upon request.

We present neural baselines based on LSTM and Transformer models. 
Our experiments show the following: (1) Transformer-based models consistently outperform Bi-LSTM-CRF-based models that have been standard for de-identification in the medical domain (\textbf{\textsc{RQ1}}). (2) Stackoverflow-related BERT representations are not more effective than regular BERT representations on Stackoverflow job postings for de-identification (\textbf{\textsc{RQ2}}). (3) MTL experiments with BERT representations and related auxiliary data sources improve our de-identification results (\textbf{\textsc{RQ3}}); the auxiliary task trained on the closer text type was the most beneficial, yet results improved with both auxiliary data sources. This shows the benefit of using multi-task learning for de-identification in job vacancy data.


\section*{Acknowledgements}

\noindent We thank the NLPnorth group for feedback on an earlier version of this paper. We would also like to thank the anonymous reviewers for their comments to improve this paper. Last, we also thank NVIDIA and the ITU High-performance Computing cluster for computing resources.  This research is supported by the Independent Research Fund Denmark (DFF) grant 9131-00019B. 
\bibliographystyle{acl_natbib}
\bibliography{nodalida2021}

\appendix

\clearpage

\clearpage
\section{Example Job Posting}\label{sec:jobpost}
\noindent\fbox{%
    \parbox{\textwidth}{%
    \small
1. \textbf{\lbrack XXX\textsubscript{Profession}\rbrack}  \\

2. \textbf{\lbrack XXX\textsubscript{Organization}\rbrack} \\

3. $<$ADDRESS$>$, $<$ADDRESS$>$, \textbf{\lbrack XXX\textsubscript{Location}\rbrack} , - , \textbf{\lbrack XXX\textsubscript{Location}\rbrack} \\

4. Date posted: 2020$-$08$-$13 \\
5. Likes: 0, Dislikes: 0, Love: 0 \\
6. Salary: SALARY \\
7. Job type: FULL\_TIME \\
8. Experience level: Mid--Level, Senior, Lead \\
9. Industry: Big Data, Cloud-Based Solutions, Enterprise Software \\
10. Company size: 501--1 \\
11. Company type: Private \\

12. Technologies: c\#, typescript, cad, 2d, 3d \\

13. Job description:\\ 
14. \textbf{\lbrack XXX\textsubscript{Organization}\rbrack} is a modern multi tenant, microservices based solution and Floor Planning is one major functional solution vertical of the \textbf{\lbrack XXX\textsubscript{Organization}\rbrack} platform. \\

15. What you’ll be doing:\\
16. As a \textbf{\lbrack XXX\textsubscript{Profession}\rbrack} for \textbf{\lbrack XXX\textsubscript{Organization}\rbrack}, you will be one of the founding members of our \textbf{\lbrack XXX\textsubscript{Location}\rbrack} based floor planning development team.\\
17. You will be in charge for development of future floor planning capabilities on the \textbf{\lbrack XXX\textsubscript{Organization}\rbrack} platform and be the software architect for the capability.\\
18. You will drive the team to improve the coding practices and boost performance.\\ 
19. You will also be a member of our \textbf{\lbrack XXX\textsubscript{Organization}\rbrack} and have a major influence on feature roadmap and technologies we use.\\

20. What you’ll bring to the table:\\
21. Solid software design and development skills and at least 5 year experience in the industry \\
22. Good understanding of CAD type of software in 2D and 3D worlds \\
23. Experience on rendering technologies, 2D/3D data models and data types \\
24. Hands-on experience in implementing CAD designers / drafting / drawing tools for on-line use \\
25. C\#, C++, TypeScript or Angular/React knowledge \\
26. Strong ambition to deliver great quality software and to continuously improve the way we do development \\
27. Good spoken and written English \\
28. Ability to work on-site in our \textbf{\lbrack XXX\textsubscript{Location}\rbrack} office, with flexible remote work possibilities \\

29. What we consider as an advantage: \\
30. Eagerness to find out and learn about the latest computer graphics technologies, and also to share your findings\\
31. Knowledge of OpenDesign components (Teigha) \\

32. What we offer you in return: \\
33. An international career and learning opportunities in a rapidly growing software company \\
34. A fun, ambitious, and committed team of smart people to work with \\
35. A respectful and professional, yet easy-going atmosphere where individual thinking is encouraged \\
36. Responsibilities in challenging projects from day one \\
37. A position where you can help retailers fight against food waste \\

38. Are you the one we’re looking for? \\
39. Apply today and become a part of our \textbf{\lbrack XXX\textsubscript{Organization}\rbrack} family! \\
40. You can apply by sending your cover letter and resume through the application form as soon as possible, but no later than 31st of August. \\
41. Please note that we will fill this position as soon as we've found the right person, so we recommend that you act quickly. \\
42. If you have questions, \textbf{\lbrack XXX\textsubscript{Name}\rbrack} (\textbf{\lbrack XXX\textsubscript{Contact}\rbrack}) from our Recruitment team is happy to answer them. \\
43. Also kindly note that we cannot process any applications through email. \\

44. Job benefits: \\
$<$cutoff$>$\\

53. Company description: \\
54. \textbf{\lbrack XXX\textsubscript{Organization}\rbrack} is a fast-growing software company developing products that help retail companies plan and operate more efficiently.\\ 
55. By accurately forecasting consumption of goods, we reduce inventory costs, increase availability and cut waste.\\
56. Helping retailers eliminate food spoilage and reduce fleet emissions from transportation has a significant environmental impact as well!
    }%
}

\end{document}